\pdfoutput=1

\documentclass[11pt]{article}

\usepackage{acl2023}

\usepackage{times}
\usepackage{latexsym}

\usepackage[T1]{fontenc}
\usepackage[utf8]{inputenc}
\usepackage{microtype}
\usepackage{inconsolata}
\usepackage{graphicx}
\usepackage{booktabs}
\usepackage{array}
\usepackage{tabularx}
\usepackage{xcolor}

\title{
RE-AD: Real-Time Requirement Adherence for Data Labeling}

\author{Siddarth Malreddy, Ishan Nigam, Akshay Arora, Nikhil Mittal, Subrat Sahu \\
  Uber AI Solutions \\
  \texttt{\{smalreddy, ishann, akshay.arora, nikhil.mittal, subrat.sahu\}@uber.com}}

\begin{document}
\maketitle
\begin{abstract}
Human-annotated data remains fundamental to training frontier Large Language Models (LLMs). However, crowd-sourced annotations often suffer from quality issues stemming from annotator misunderstanding or lack of engagement. To address this, we introduce a real-time requirement adherence (\textbf{RE-AD}) framework  that leverages LLMs to proactively validate labeling quality. Our methodology involves decomposing Standard Operating Procedures (SOPs) into atomic rules via self-reflection, categorizing them by complexity, and applying tiered validation strategies. Evaluated on a synthetic benchmark, the system achieved an F1 score of 0.749. Furthermore, production deployment resulted in annotators accepting and fixing 82\% of the errors flagged by the framework. We include ablation studies to demonstrate the impact of our core design decisions.

\end{abstract}

\section{Introduction}

High-quality annotated data is the bedrock of supervised machine learning \citep{pustejovsky2012natural}, yet maintaining rigorous annotation quality remains a persistent bottleneck. Traditional quality assurance (QA) methodologies predominantly rely on post-hoc evaluations, such as inter-annotator agreement metrics \citep{artstein2008inter} or expert spot-checks \citep{snow2008cheap}. These approaches are inherently \textit{reactive}: errors are identified only after the data collection cycle is complete, necessitating costly and time-consuming rework. 

The challenge is exacerbated in specialized domains where annotators must adhere to multifaceted Standard Operating Procedures (SOPs). These guidelines often encompass a heterogeneous range of constraints, from rigid structural formatting 
to nuanced, high-level linguistic properties 
. In such settings, the cognitive load on human annotators increases the likelihood of "requirement drift," where subtle guidelines are overlooked during the generation process.

\begin{figure}[t]
\centering
\includegraphics[width=\columnwidth]{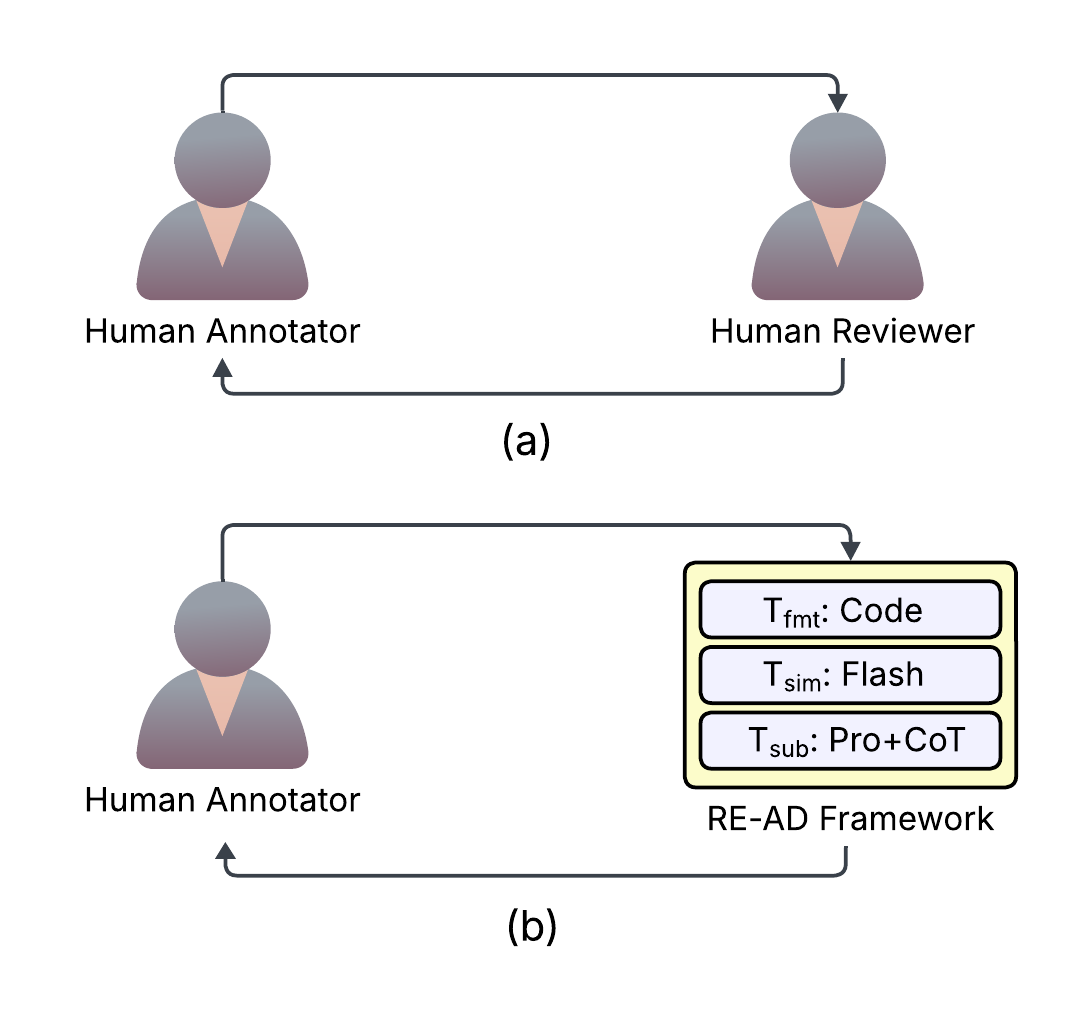}
\vspace{-13mm}
\caption{
\textbf{(a) Traditional Data Labeling Pipeline:} Conventional QA relies on post-hoc audits which identify errors only after the data collection cycle is complete and necessitate time-consuming rework. This effectively decouples quality control from the point of data origin and traps annotators in a reactive cycle.\\
\textbf{(b) RE-AD Framework:} Transitions quality control to a proactive assistance model by atomizing guidelines into three complexity-aware tiers ($T_{fmt}$, $T_{sim}$, $T_{sub}$) to provide real-time feedback within 2.3s. Production results show an 82\% acceptance and fix rate, effectively preventing requirement drift before the final audit.}
\label{fig:validation_design}
\end{figure}

Recent advances in Large Language Models (LLMs) have demonstrated remarkable proficiency in instruction following \citep{ouyang2022instructgpt} and complex reasoning \citep{xie2024llmannotationsurvey}. While significant research has focused on using LLMs as primary annotators \citep{gilardi2023chatgpt, he2024annollm} or as "post-hoc judges" \citep{zheng2023llmjudge}, their potential for \textit{proactive} validation within the human annotation loop remains under-explored. 

In this work, we introduce the \textbf{RE-AD} (\textbf{RE}quirement \textbf{AD}herence) framework for real-time requirement adherence that transitions quality control from a reactive audit to a proactive assistance model (Figure~\ref{fig:validation_design}). Our system automatically atomizes complex guidelines into a hierarchy of verifiable constraints and employs a complexity-aware routing architecture to validate human input as it is generated. This ensures that annotators receive immediate feedback, reducing the error rate before the data reaches the final audit stage. The system consists of two primary components: an offline constraint atomization pipeline (Figure~\ref{fig:atomization}) and an online, complexity-aware validation engine (Figure~\ref{fig:validation}). Our contributions are as follows:
\begin{itemize}
    \item \textbf{Recursive Constraint Atomization:} A self-reflective methodology for decomposing unstructured guidelines into verifiable rules categorized into a three-tier complexity hierarchy.\vspace{-3mm}
    \item \textbf{Complexity-Aware Validation Pipeline:} A production-ready architecture utilizing parallel per-rule validation, prefix caching, and adaptive model routing (spanning deterministic code to high-reasoning LLMs) to balance evaluation accuracy with latency.\vspace{-3mm}
    \item \textbf{Systematic Evaluation:} An empirical analysis using a novel synthetic benchmark with controlled violation injection, alongside an ablation study comparing atomic per-rule validation against holistic batch processing.\vspace{-3mm}
    \item \textbf{Impact Analysis:} Results from a large-scale production deployment show an 82\% acceptance rate of errors flagged by the framework.
\end{itemize}

\section{Related Work}



\paragraph{Annotation Quality Assurance} 
Traditional methodologies for maintaining annotation quality have historically relied on post-hoc evaluations. These include inter-annotator agreement (IAA) metrics such as Cohen's Kappa or Krippendorff's Alpha \citep{artstein2008inter}, expert gold-standard comparisons, and redundant labeling strategies \citep{snow2008cheap}. While frameworks such as active learning \citep{settles2009active} and human-in-the-loop (HITL) systems \citep{mosqueira2023humanhitl} have improved data efficiency, they remain fundamentally \textit{reactive}, identifying errors only after a labeling cycle is complete.

\paragraph{LLM-as-a-Judge} 
The emergence of LLMs has shifted the landscape of evaluation. The "LLM-as-a-judge" paradigm \citep{zheng2023llmjudge} has demonstrated that high-parameter models can effectively validate outputs against complex, structured requirements. Recent work has explored the reliability of these automated judges using techniques like self-reflection \citep{ji2023llmannotation, madaan2023selfrefine} and self-consistency \citep{wang2022selfconsistency}. Our work builds on these foundations while applying these techniques to \textit{real-time} human-in-the-loop validation rather than static model evaluation.

\paragraph{Constrained Generation and Verification} 
The task of verifying adherence to Standard Operating Procedures (SOPs) is a form of constrained generation. While much literature focuses on \textit{decoding} with constraints \citep{lu2020constrained}, less focus is placed on the \textit{verification} of arbitrary natural language guidelines. Recent surveys on LLM evaluation \citep{gu2024survey} highlight the difficulty of evaluating "long-tail" subjective constraints. Our tiered architecture addresses this by matching the verification method's complexity to the constraint type, a strategy rarely explored in existing literature on data labeling workflows.

\section{Methodology}

%
%

\textbf{RE-AD} is designed to provide annotators with real-time feedback on complex linguistic constraints. The system comprises of two distinct stages: (1) \textbf{Constraint Atomization}, where unstructured guidelines are atomized into a machine-verifiable schema (Sec. 3.1), and (2) \textbf{Tiered Validation Architecture}, where rules are executed via a complexity-aware, tiered routing logic (Sec. 3.2).

\subsection{Constraint Atomization}

\begin{figure*}[ht]
    \centering
    \includegraphics[width=1.8\columnwidth]{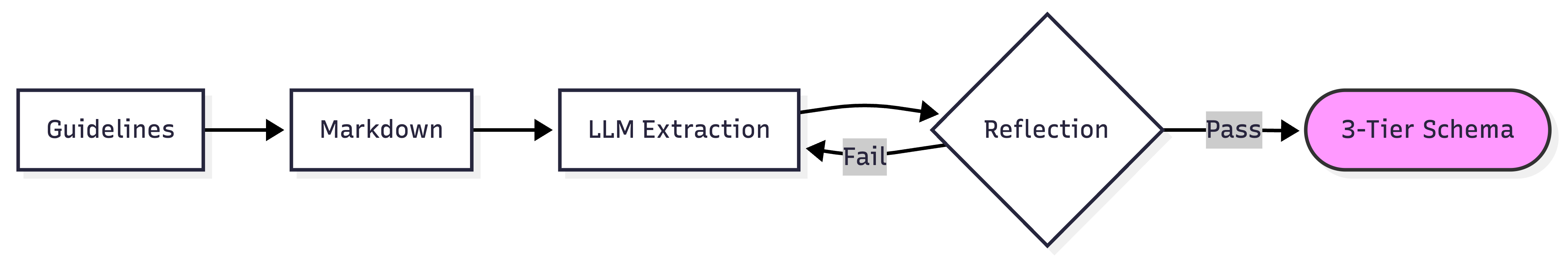}
    \caption{Phase 1: Recursive Constraint Atomization. The pipeline illustrates the offline transformation of unstructured guidelines into an orthogonal, three-tier rule schema via a self-reflective LLM loop, ensuring each constraint is atomic and verifiable.}
    \label{fig:atomization}
\end{figure*}

%

The first phase involves converting unstructured guidelines into a machine-verifiable schema. Building on recent findings in LLM-based natural language understanding \citep{xie2024llmannotationsurvey}, we employ a \textit{self-reflective decomposition} strategy to extract atomic rules, as illustrated in Figure~\ref{fig:atomization}.

To normalize varying client requirements, we first convert unstructured SOPs (e.g., PDF, DOCX) into structured Markdown to preserve hierarchical relationships. Rather than a single-pass extraction, which is prone to information loss and hallucination, an LLM agent iteratively identifies candidate rules. For each candidate, a secondary reflection step \citep{ji2023llmannotation} verifies \textbf{atomicity} (ensuring the rule contains only one logical constraint) and \textbf{orthogonality} (ensuring no semantic overlap with existing rules). Rules failing these criteria are recursively decomposed or merged. Empirically, we found the atomization process to be stable, with less than 5\% variance in rule composition across multiple runs of the same SOP.

Validated rules are mapped to a three-tier hierarchy. Table~\ref{tab:example_rules} illustrates representative rule-sets extracted from a conversation generation SOP.



\begin{table}[ht]\centering
\small
\begin{tabularx}{\columnwidth}{@{}l X@{}}
\toprule
\textbf{Tier} & \textbf{Linguistic Rule / Constraint} \\
\midrule
\textbf{Formatting} & Each message between 20--150 words. \\
($T_{fmt}$)         & Exact 10 turns (5 per participant). \\
\midrule
\textbf{Simple}     & No profane words (e.g., ``hell''). \\
($T_{sim}$)         & First message must include a greeting. \\
\midrule
\textbf{Subjective} & Professional tone; avoid slang. \\
($T_{sub}$)         & Indian context via names/context. \\
\bottomrule
\end{tabularx}
\caption{Examples of atomic rules extracted from an industrial SOP using our recursive reflection pipeline. Rules are categorized into a three-tier complexity hierarchy ($T_{fmt}$ through $T_{sub}$) which dictates the downstream validation strategy: deterministic scripts for $T_{fmt}$ and tiered LLM routing for $T_{sim}$ and $T_{sub}$.}
\label{tab:example_rules}
\end{table}

%
%
%

\subsection{Tiered Validation Architecture}

\begin{figure*}[ht]
    \centering
    \includegraphics[width=1.8\columnwidth]{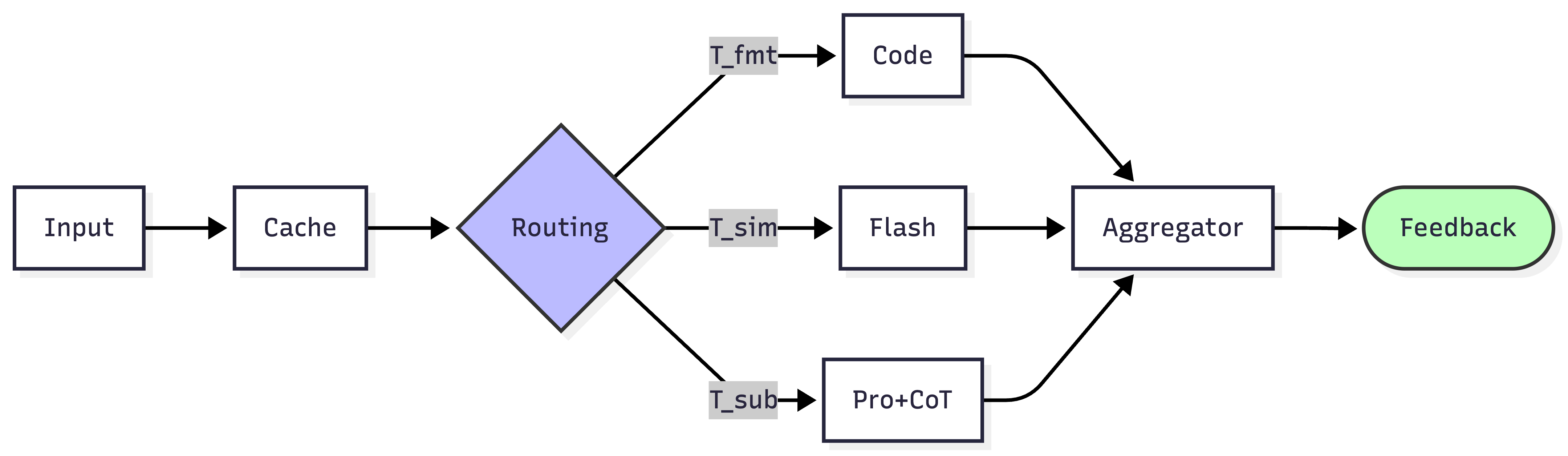}
    \caption{Phase 2: Tiered Parallel Validation Architecture. The online engine utilizes a prefix caching layer to minimize latency and routes human input through complexity-aware validators ($T_{fmt}$, $T_{sim}$, $T_{sub}$) in parallel. Results are aggregated to provide proactive feedback within the annotation interface.}
    \label{fig:validation}
\end{figure*}

The core of the \textbf{RE-AD} framework is a parallel, complexity-aware validation engine (Figure~\ref{fig:validation}). Rather than evaluating a submission against a monolithic prompt, which incurs high latency and potential ``lost-in-the-middle'' issues \citep{liu2024lostinthemiddle}, our system executes validation in parallel across the hierarchy defined in Table~\ref{tab:example_rules}. This architecture leverages the strong instruction-following capabilities of modern LLMs \citep{ouyang2022instructgpt} to provide granular, per-rule feedback.

%
%
%

\paragraph{Deterministic Routing ($T_{fmt}$)} 
Constraints classified as formatting (e.g., word counts or turn limits) bypass LLM inference entirely. Building on best practices for hybrid annotation systems \citep{xie2024llmannotationsurvey}, we utilize 
Python-based regex and string-parsing validators ensuring 100\% precision for structural requirements while reducing the total token consumption of the validation cycle.

\paragraph{Lightweight Inference ($T_{sim}$)}
For simple lexical constraints (e.g., keywords or greetings), we route rules to a high-throughput small-parameter model (Gemini 3 Flash). These checks rely on zero-shot prompting, as the model's primary task is pattern matching and not deep semantic reasoning.

\paragraph{Reasoning-Heavy Routing ($T_{sub}$)} 
Subjective rules (e.g., tone adherence) are routed to a high-capacity model (Gemini 3 Pro). To ensure reliability in these ``gray-area'' judgments, we implement \textbf{Chain-of-Thought (CoT)} prompting, where we prompt the model to generate a brief rationale before outputting a binary pass/fail \citep{wei2022chainofthought}.

\paragraph{Low-Latency Optimization} 
To meet the requirements of a real-time annotation interface, we utilize \textbf{prefix caching}. By caching the system prompt and the atomized SOP rules, the system only processes the newly generated turn in each validation cycle. This reduces Time-To-First-Token (TTFT) and maintains a seamless feedback loop for the human annotator.

\section{Evaluation}

We evaluate the \textbf{RE-AD} framework through a two-pronged approach focusing on both controlled accuracy and production-scale utility. First, we describe the construction of \textbf{RE-AD-Eval}, a synthetic benchmark designed to quantify validation reliability across our three complexity tiers ($T_{fmt}$, $T_{sim}$, $T_{sub}$) and varying annotator skill levels (Section 4.1). Second, we present a performance analysis of our tiered routing architecture, including an ablation study that highlights the trade-offs between validation granularity and system latency (Section 4.2). Finally, we report results from a large-scale production deployment to measure the framework's impact on reducing audit overhead in industrial labeling workflows (Section 4.3).

\subsection{Synthetic Benchmark: RE-AD-Eval}

We construct a synthetic benchmark with known ground-truth labels to enable controlled evaluation. The benchmark is generated through a five-stage pipeline using OpenAI models for synthesis, while our system uses Gemini models during evaluation to prevent model bias.

\subsubsection{SOP Generation} 
An LLM generates a 20-rule SOP for a conversational data collection task, spanning three complexity tiers: formatting ($T_{fmt}$, 8 rules), simple ($T_{sim}$, 5 rules), and subjective ($T_{sub}$, 7 rules). The SOP is designed to cover diverse rule types representative of real-world industrial requirements.

\subsubsection{Scenario Diversification} 
To ensure broad conversational coverage and prevent rule evaluation from being confounded with topic familiarity, we generate diverse scenario tuples. These include varying participant names, settings, and conversational objectives, ensuring the validator's performance is tested across heterogeneous contexts.

\subsubsection{Clean Conversation Generation} 
For each scenario, we generate a baseline "clean" conversation that satisfies all 20 rules. These conversations serve as the gold-standard foundation for our controlled violation injection process.

\subsubsection{Error Injection with Verify-and-Repair} 
Three annotator personas (Expert, Intermediate, Novice) are defined with tier-specific error probabilities: 5--25\% for $T_{fmt}$, 8--35\% for $T_{sim}$, and 12--50\% for $T_{sub}$. Each sample receives exactly one injected violation to enable precise per-rule attribution. A verify-and-repair loop (up to 3 iterations) confirms that the intended violation is present and no extraneous errors were introduced. While injecting a single violation allows for precise attribution, we acknowledge this is a simplification of real-world scenarios where annotators may commit compound or cascading errors. During evaluations, we check if the errors injected are identified by the framework.

\subsubsection{Hybrid Ground Truth} 
Ground truth is established via a hybrid strategy: deterministic code-based checks for $T_{fmt}$ and independent LLM verification for $T_{sim}$ and $T_{sub}$. This ensures reliable ground truth across both structural and semantic complexity tiers.

\subsection{Results}

We conduct a comprehensive performance analysis focusing on accuracy, latency, and the specific trade-offs inherent in our tiered architecture. We first report the validation accuracy across different annotator skill levels and complexity tiers (Section 4.2.1), followed by a qualitative error analysis of subjective constraints (Section 4.2.2). Finally, we present an ablation study comparing our tiered parallel approach against monolithic batch processing baselines (Section 4.2.3). All of the metrics are computed using micro-average across samples.

\subsubsection{Validation Accuracy}


As reported in Tables~\ref{tab:prod_agent} and~\ref{tab:prod_tier}, the \textbf{RE-AD} framework demonstrates high reliability across varying annotator expertise and rule complexities. The system maintains a stable F1-score of 0.74--0.77 across simulated skill levels, indicating that the validation logic is robust to the types of errors characteristic of different annotator profiles. Note that the $T_{fmt}$ validator used in the benchmark is not the same script used for ground truth and is re-generated using an LLM. When disaggregated by complexity tier, performance remains highest for formatting constraints ($T_{fmt}$ F1: 1.000) and degrades gracefully as the requirement becomes more semantic ($T_{sub}$ F1: 0.551).

\begin{table}[t]
\centering
\small
\begin{tabular}{@{}l c c c@{}}
\toprule
\textbf{Skill Level} & \textbf{Prec.} & \textbf{Recall} & \textbf{F1} \\
\midrule
   Expert & 0.665 & 0.907 & 0.767 \\
  Intermediate       & 0.621 & 0.915 & 0.740 \\
  Novice       & 0.617 & 0.913 & 0.737 \\
\bottomrule
\end{tabular}
\caption{Validation performance on the RE-AD-Eval benchmark across simulated annotator skill levels. The framework maintains a consistent F1-score ($\sim$0.75) despite varying error distributions in the input. This suggests that the validation logic is resilient to fluctuations in human labeling quality and remains effective for both expert-level refinement and novice-level intervention.}
\label{tab:prod_agent}
\end{table}

\subsubsection{Error Analysis} 

Errors within the subjective tier ($T_{sub}$) frequently arise from interpretive ambiguity rather than systemic model failure. Constraints such as ``maintain a professional tone'' admit a range of reasonable interpretations. In several cases, the validator's judgment diverged from the ground truth due to cultural reasoning or boundary cases where even expert annotators often disagree. This mirrors established findings in the ``LLM-as-judge'' literature \citep{zheng2023llmjudge}. Consequently, $T_{sub}$ validations are best utilized as assistive signals that surface suggestions for human review rather than as automated rejection gates.

\subsubsection{Ablation: Per-Rule vs.\ Batch}

%
%
%

Table~\ref{tab:ablation} compares our parallel per-rule architecture against monolithic baselines. The per-rule strategy demonstrates a clear advantage in both efficiency and structural accuracy. The best batch configuration (Gemini 3 Pro) requires over 16$\times$ the wall-clock time compared to \textbf{RE-AD} (36.8s vs. 2.3s), rendering it unsuitable for real-time applications.

Furthermore, batch models struggle with $T_{fmt}$ constraints, often failing to verify structural counts within a large context (F1 as low as 0.471). RE-AD benefits from using deterministic code for these checks. While batch processing shows a performance benefit on $T_{sim}$ and $T_{sub}$ tasks, likely due to interdependencies between the rules, the latency-to-accuracy trade-off strongly favors the tiered, parallelized approach for in-tool assistance. However, the higher accuracy of Batch models on $T_{sub}$ ($0.612$ vs. $0.551$) suggests that atomization may strip away holistic context necessary for some subjective judgments. RE-AD's higher Total F1 is largely driven by its superior performance on rote $T_{fmt}$ tasks.

\begin{table}[t!]
\centering
\small
\begin{tabular}{@{}l c c c@{}}
\toprule
\textbf{Rule Tier} & \textbf{Prec.} & \textbf{Recall} & \textbf{F1} \\
\midrule
  Formatting ($T_{fmt}$)    & 1.000 & 1.000 & 1.000 \\
  Simple ($T_{sim}$) & 0.908 & 0.824 & 0.864 \\
  Subjective ($T_{sub}$)    & 0.402 & 0.876 & 0.551 \\
\bottomrule
\end{tabular}
\caption{System accuracy disaggregated by complexity tier. The results quantify a clear ``gradient of reliability'': deterministic formatting constraints ($T_{fmt}$) are at human-level parity, while subjective semantic assessments ($T_{sub}$) exhibit expected degradation. This variance is driven by inherent linguistic ambiguity and highlights the role of the system as an assistive signal for gray-area requirements.}
\label{tab:prod_tier}
\end{table}

\begin{table*}[t]
\centering
\small
\begin{tabular}{@{}l c c c c c c c@{}}
\toprule
\textbf{Configuration} & \textbf{p50 Latency(s)} & \textbf{Total F1} & \textbf{Fmt F1} & \textbf{Sim F1} & \textbf{Subj F1} \\
\midrule
\textbf{RE-AD (3-Flash + 3-Pro)} & 2.278 & \textbf{0.749} & \textbf{1.000} & 0.864 & 0.551 \\
\midrule
Batch (3-Flash) & 5.823 & 0.628 & 0.471 & 0.890 & 0.590 \\
Batch (3-Pro) & 36.853 & 0.742 & 0.756 & \textbf{0.892} & \textbf{0.612} \\
\bottomrule
\end{tabular}
\caption{Ablation analysis comparing our proposed parallel per-rule architecture (RE-AD) against monolithic batch processing. RE-AD provides an order-of-magnitude reduction in median latency while ensuring structural accuracy. Notably, while batch models leverage holistic context for slight gains in $T_{sub}$ accuracy, they suffer from significant ``lost-in-the-middle'' degradation on formatting tasks.}
\label{tab:ablation}
\end{table*}

\subsection{Production Deployment} 


To evaluate the practical utility of our framework beyond controlled benchmarks, we integrated \textbf{RE-AD} into a high-volume industrial labeling workflow. This section reports the operational impact observed during a month-long deployment, focusing on the reduction of post-labeling audit overhead and the framework's viability for real-time human-in-the-loop assistance.

Over a month-long deployment period, the per-rule variant resulted in annotators accepting and fixing \textbf{82\%} of the errors flagged by the framework. The other 18\% were false positives that were dismissed by the human annotators. This significant operational impact confirms that proactive, real-time feedback effectively mitigates requirement drift and reduces the total overhead of reactive human auditing. This rate likely reflects a combination of the tool's utility, the natural learning curve of the annotators, and concurrent UI optimizations. To prevent alert fatigue, especially given $T_{sub}$'s low precision, these flags are presented as non-blocking suggestions rather than mandatory gates. At the same time, we see scope for future improvement.


\section{Discussion and Conclusion}

Our results quantify a clear gradient of reliability across the complexity tiers: while the system reaches human parity on formatting constraints 
, performance degrades on subjective assessments 
. This divergence is consistent with broader findings in the LLM-as-judge literature regarding the difficulty of evaluating open-ended semantic properties \citep{zheng2023llmjudge, zhang2023llmeval}. 

The comparison between per-rule and batch evaluation highlights a critical trade-off: \textit{granularity vs. context}. Isolated evaluation (per-rule) prevents ``lost-in-the-middle'' performance degradation and enables precise error feedback, which is vital for Simple ($T_{sim}$) checks. However, subjective judgments ($T_{sub}$) appear to benefit from the holistic context provided in batch mode, suggesting that reasoning-heavy rules are inherently non-atomic.

In this work, we introduced \textbf{RE-AD}, a framework for real-time linguistic constraint adherence. By atomizing guidelines and deploying a tiered validation engine, we achieved significant operational impact, with an 82\% acceptance and fix rate for flagged errors. We hypothesize that this high acceptance rate is a result of the framework's real-time performance; with a median latency of \textbf{2.3s}. The system provides feedback while the annotator is engaged, preventing requirement drift before the data reaches post-hoc auditing. 
While tested on English conversations, our three-tier taxonomy (formatting, simple-lexical, and subjective-semantic) is designed to be universal across specialized domains like medicine or law, which share similar hierarchical guideline structures.

\section*{Limitations}


\textbf{RE-AD} demonstrates significant operational utility, though several limitations remain. Our evaluation is currently restricted to English-language tasks; the generalizability of our complexity tiers to morphologically rich or low-resource languages remains to be tested.Further research is also required to assess RE-AD's robustness in noisy, multilingual, or domain-shifted workflows where the `ground truth' of an SOP may be more fluid. While \textbf{RE-AD} enables controlled experimentation, it may not capture the full range of stochastic or edge-case errors committed by human annotators in the wild. 

The significant performance degradation for $T_{sub}$ achieving an F1 of $0.551$ underscores the inherent difficulty in automating the validation of subjective linguistic requirements like tone and cultural nuance. Finally, our reported 82\% acceptance rate in quality audits is an operational estimate from a production environment. We acknowledge that confounding variables such as concurrent UI improvements, the natural learning curve of the annotator pool, and shifts in project complexity may contribute to this observed rate. Future studies are required to isolate the specific causal impact of real-time feedback on long-term data quality.

\section*{Ethics Statement}

\textbf{RE-AD} involves the use of LLMs to assist in the quality assurance of human-annotated data. We identify three primary ethical considerations:
\vspace{1.5mm}
\paragraph{Impact on Human Annotators} 
Our primary goal is to transition quality assurance from a reactive, punitive audit model to a proactive, assistive one. While our results show an 82\% acceptance rate, we emphasize that the framework is designed to augment human expertise, not to replace human labor. The assistive loop is intended to reduce the cumulative cognitive load associated with intensive post-hoc data correction cycles.
\vspace{1.5mm}
\paragraph{Data Privacy and Security} 
The production data utilized for impact analysis (Sec. 4.3) was anonymized to remove all Personally Identifiable Information (PII) before being processed by LLM validators. Our framework utilizes enterprise-grade API endpoints with strict data-retention policies to ensure that proprietary client guidelines and annotator submissions remain secure.
\vspace{1.5mm}
\paragraph{Synthetic Data and Bias} 
The benchmarking data was synthetically generated. We acknowledge that LLMs may carry inherent biases related to cultural norms, tone, and linguistic variety. To mitigate this, our three-tier complexity hierarchy includes a specific ``Subjective'' tier where human review is prioritized. We explicitly categorize $T_{sub}$ judgments as assistive signals rather than automated gates to ensure the system remains a guide rather than a definitive arbiter of linguistic nuance.

\newpage

\bibliographystyle{acl_natbib}
\bibliography{custom}

\begin{thebibliography}{19}
\providecommand{\natexlab}[1]{#1}

\bibitem[{Artstein and Poesio(2008)}]{artstein2008inter}
Ron Artstein and Massimo Poesio. 2008.
\newblock Inter-coder agreement for computational linguistics.
\newblock \emph{Computational Linguistics}, 34(4):555--596.

\bibitem[{Gilardi et~al.(2023)Gilardi, Alizadeh, and
  Kubli}]{gilardi2023chatgpt}
Fabrizio Gilardi, Meysam Alizadeh, and Maoel Kubli. 2023.
\newblock {ChatGPT} outperforms crowd-workers for text-annotation tasks.
\newblock \emph{Proceedings of the National Academy of Sciences}, 120(30).

\bibitem[{He et~al.(2024)He, Lin, Gong, Jin, Zhang, Lin, Jiao, Yiu, Duan, and
  Chen}]{he2024annollm}
Xingwei He, Zhenghao Lin, Yeyun Gong, Alex Jin, Hang Zhang, Chen Lin, Jian
  Jiao, Siu~Ming Yiu, Nan Duan, and Weizhu Chen. 2024.
\newblock {AnnoLLM}: Making large language models to be better crowdsourced
  annotators.
\newblock In \emph{Proceedings of the 2024 Conference of the North American
  Chapter of the Association for Computational Linguistics}, pages 4196--4212.

\bibitem[{Ji et~al.(2023)Ji, Lee, Frieske, Yu, Su, Xu, Ishii, Bang, Madotto,
  and Fung}]{ji2023llmannotation}
Ziwei Ji, Nayeon Lee, Rita Frieske, Tiezheng Yu, Dan Su, Yan Xu, Etsuko Ishii,
  Yejin Bang, Andrea Madotto, and Pascale Fung. 2023.
\newblock Towards mitigating {LLM} hallucination via self reflection.
\newblock In \emph{Findings of the Association for Computational Linguistics:
  EMNLP 2023}, pages 1827--1843.

\bibitem[{Li et~al.(2023)Li, Shi, Ziems, Rber, and Yang}]{li2023coannotating}
Minzhi Li, Taiwei Shi, Caleb Ziems, Min-Yen Rber, and Diyi Yang. 2023.
\newblock Co{A}nnotating: Uncertainty-guided work allocation between human and
  large language models for data annotation.
\newblock In \emph{Proceedings of the 2023 Conference on Empirical Methods in
  Natural Language Processing}, pages 1487--1505.

\bibitem[{Madaan et~al.(2023)Madaan, Tandon, Gupta, Hallinan, Gao, Wiegreffe,
  Alon, Dziri, Prabhumoye, Yang et~al.}]{madaan2023selfrefine}
Aman Madaan, Niket Tandon, Prakhar Gupta, Skyler Hallinan, Luyu Gao, Sarah
  Wiegreffe, Uri Alon, Nouha Dziri, Shrimai Prabhumoye, Yiming Yang, and 1
  others. 2023.
\newblock Self-refine: Iterative refinement with self-feedback.
\newblock In \emph{Advances in Neural Information Processing Systems},
  volume~36, pages 46534--46594.

\bibitem[{Mosqueira-Rey et~al.(2023)Mosqueira-Rey, Hernandez-Pereira,
  Alonso-Rios, Bobes-Basc{\'a}n, and Fernandez-Leal}]{mosqueira2023humanhitl}
Eduardo Mosqueira-Rey, Elena Hernandez-Pereira, David Alonso-Rios, Jos{\'e}
  Bobes-Basc{\'a}n, and Angeles Fernandez-Leal. 2023.
\newblock Human-in-the-loop machine learning: a state of the art.
\newblock \emph{Artificial Intelligence Review}, 56(4):3005--3054.

\bibitem[{Ouyang et~al.(2022)Ouyang, Wu, Jiang, Almeida, Wainwright, Mishkin,
  Zhang, Agarwal, Slama, Ray et~al.}]{ouyang2022instructgpt}
Long Ouyang, Jeffrey Wu, Xu~Jiang, Diogo Almeida, Carroll Wainwright, Pamela
  Mishkin, Chong Zhang, Sandhini Agarwal, Katarina Slama, Alex Ray, and 1
  others. 2022.
\newblock Training language models to follow instructions with human feedback.
\newblock In \emph{Advances in Neural Information Processing Systems},
  volume~35, pages 27730--27744.

\bibitem[{Pustejovsky and Stubbs(2012)}]{pustejovsky2012natural}
James Pustejovsky and Amber Stubbs. 2012.
\newblock \emph{Natural Language Annotation for Machine Learning}.
\newblock O'Reilly Media.

\bibitem[{Ramnath et~al.(2024)Ramnath, Zhou, Chen, Xie, Ding, Zhang, Chen, and
  Lam}]{ramnath2024promptsurvey}
Kiran Ramnath, Kang Zhou, Yulin Chen, Yibo Xie, Kaikai Ding, Yiyou Zhang,
  Zeyuan Chen, and Wai Lam. 2024.
\newblock A systematic survey of prompt engineering in large language models:
  Techniques and applications.
\newblock In \emph{Findings of the Association for Computational Linguistics:
  ACL 2024}, pages 10672--10685.

\bibitem[{Settles(2009)}]{settles2009active}
Burr Settles. 2009.
\newblock Active learning literature survey.
\newblock \emph{Computer Sciences Technical Report 1648, University of
  Wisconsin--Madison}.

\bibitem[{Snow et~al.(2008)Snow, O'Connor, Jurafsky, and Ng}]{snow2008cheap}
Rion Snow, Brendan O'Connor, Daniel Jurafsky, and Andrew~Y. Ng. 2008.
\newblock Cheap and fast --- but is it good? {E}valuating non-expert
  annotations for natural language tasks.
\newblock In \emph{Proceedings of the 2008 Conference on Empirical Methods in
  Natural Language Processing}, pages 254--263.

\bibitem[{Wang et~al.(2023)Wang, Wei, Schuurmans, Le, Chi, Sharan, Chowdhery,
  and Zhou}]{wang2022selfconsistency}
Xuezhi Wang, Jason Wei, Dale Schuurmans, Quoc Le, Ed~Chi, Narang Sharan,
  Aakanksha Chowdhery, and Denny Zhou. 2023.
\newblock Self-consistency improves chain of thought reasoning in language
  models.
\newblock In \emph{International Conference on Learning Representations}.

\bibitem[{Wei et~al.(2022)Wei, Wang, Schuurmans, Bosma, Ichter, Xia, Chi, Le,
  and Zhou}]{wei2022chainofthought}
Jason Wei, Xuezhi Wang, Dale Schuurmans, Maarten Bosma, Brian Ichter, Fei Xia,
  Ed~Chi, Quoc Le, and Denny Zhou. 2022.
\newblock Chain-of-thought prompting elicits reasoning in large language
  models.
\newblock In \emph{Advances in Neural Information Processing Systems},
  volume~35, pages 24824--24837.

\bibitem[{Xie et~al.(2024)Xie, Li, Zhang, Gao, Zhang, Zhang, Chen, and
  Liu}]{xie2024llmannotationsurvey}
Zhen Xie, Yao Li, Yuxiang Zhang, Minglai Gao, Wenwen Zhang, Rui Zhang, Weiran
  Chen, and Qing Liu. 2024.
\newblock Large language models for data annotation and synthesis: A survey.
\newblock In \emph{Proceedings of the 2024 Conference on Empirical Methods in
  Natural Language Processing}, pages 927--958.

\bibitem[{Yan et~al.(2024)Yan, Sun, An, Qian, Li, Qiu, and
  Huang}]{yan2024mirror}
Hanqi Yan, Qinglin Sun, Yuezihan An, Ziang Qian, Hongyu Li, Xipeng Qiu, and
  Xuanjing Huang. 2024.
\newblock Mirror: A multiple-perspective self-reflection method for
  knowledge-rich reasoning.
\newblock In \emph{Proceedings of the 62nd Annual Meeting of the Association
  for Computational Linguistics}, pages 6167--6181.

\bibitem[{Zhang et~al.(2023)Zhang, Yu, Yu, Lv, Liu, Huang, Xu, and
  Li}]{zhang2023llmeval}
Xinghua Zhang, Bowen Yu, Haiyang Yu, Yangyu Lv, Tingwen Liu, Fei Huang, Hongbo
  Xu, and Yongbin Li. 2023.
\newblock Wider and deeper {LLM} networks are fairer {LLM} evaluators.
\newblock \emph{arXiv preprint arXiv:2308.01862}.

\bibitem[{Zheng et~al.(2023)Zheng, Chiang, Sheng, Zhuang, Wu, Zhuang, Lin, Li,
  Li, Xing et~al.}]{zheng2023llmjudge}
Lianmin Zheng, Wei-Lin Chiang, Ying Sheng, Siyuan Zhuang, Zhanghao Wu, Yonghao
  Zhuang, Zi~Lin, Zhuohan Li, Dacheng Li, Eric Xing, and 1 others. 2023.
\newblock Judging {LLM}-as-a-judge with {MT}-bench and chatbot arena.
\newblock In \emph{Advances in Neural Information Processing Systems},
  volume~36, pages 46595--46623.

\bibitem[{Zhou et~al.(2023)Zhou, Muresanu, Han, Paster, Pitis, Chan, and
  Ba}]{zhou2022autoprompt}
Yongchao Zhou, Andrei~Ioan Muresanu, Ziwen Han, Keiran Paster, Silviu Pitis,
  Harris Chan, and Jimmy Ba. 2023.
\newblock Large language models are human-level prompt engineers.
\newblock In \emph{International Conference on Learning Representations}.

\end{thebibliography}

\end{document}